\documentclass{article}

\usepackage{microtype}
\usepackage{graphicx}
\usepackage{subfigure}
\usepackage{booktabs} 

\usepackage{hyperref}
\usepackage{multirow}

\usepackage[accepted]{multi}

\usepackage{amsmath}
\usepackage{amssymb}
\usepackage{mathtools}
\usepackage{amsthm}

\usepackage[capitalize,noabbrev]{cleveref}

\theoremstyle{plain}

\theoremstyle{definition}

\theoremstyle{remark}

\usepackage[textsize=tiny]{todonotes}

\icmltitlerunning{Training Sparse Mixture Of Experts Text Embedding Models}

\begin{document}

\twocolumn[
\icmltitle{Training Sparse Mixture Of Experts Text Embedding Models}



\icmlsetsymbol{equal}{*}

\begin{icmlauthorlist}
\icmlauthor{Zach Nussbaum}{equal,nomic}
\icmlauthor{Brandon Duderstadt}{nomic}
\end{icmlauthorlist}

\icmlaffiliation{nomic}{Nomic AI, New York, NY, USA}

\icmlcorrespondingauthor{Zach Nussbaum}{zach@nomic.ai}

\icmlkeywords{Machine Learning, ICML}

\vskip 0.3in
]



\printAffiliationsAndNotice{}  

\begin{abstract}
Transformer-based text embedding models have improved their performance on benchmarks like MIRACL and BEIR by increasing their parameter counts.
However, this scaling approach introduces significant deployment challenges, including increased inference latency and memory usage.
These challenges are particularly severe in retrieval-augmented generation (RAG) applications, where large models' increased memory requirements constrain dataset ingestion capacity, and their higher latency directly impacts query-time performance.
While causal language models have addressed similar efficiency challenges using Mixture of Experts (MoE) architectures, this approach hasn't been successfully adapted to the general text embedding setting.
In this paper, we introduce Nomic Embed v2, the first general purpose MoE text embedding model.
Our model outperforms models in the same parameter class on both monolingual and multilingual benchmarks while also maintaining competitive performance with models twice its size.
We open-source all code, models, and evaluation data to ensure full reproducibility of our training pipeline at
\href{https://github.com/nomic-ai/contrastors}{https://github.com/nomic-ai/contrastors}.
\end{abstract}

\section{Introduction}
Transformer-based biencoders are the standard architecture for training dense sentence embedding models for text retrieval \citep{reimers2019sentencebert}.
In the monolingual setting, these models are trained on curated internet-scale data \citep{wang2024textembeddingsweaklysupervisedcontrastive, xiao2024cpackpackedresourcesgeneral, günther2023jina, nussbaum2024nomicembedtrainingreproducible, li2023general}, and sometimes augmented with task-specific instructions \citep{su2023embedder}.
While models like mE5 \citep{wang2024multilinguale5textembeddings}, BGE-M3 \citep{chen2024bgem3embeddingmultilingualmultifunctionality}, mGTE \citep{zhang2024mgtegeneralizedlongcontexttext}, and Jina V3 \citep{günther2024jina} make strides towards a unified embedding space across languages, they underperform their parameter-equivalent monolingual counterparts on English benchmarks.
Multilingual models primarily close this performance gap by increasing their parameter counts, often through the use of large, pretrained multilingual Language Models fine-tuned for retrieval applications \citep{jiang2023mistral, lee2024nvembedimprovedtechniquestraining}.

The large size of multilingual embedding models creates significant deployment challenges. Their substantial memory requirements and increased inference latency particularly impact retrieval-augmented generation (RAG) applications, where they constrain both dataset ingestion capacity and query-time performance.

While causal language models have addressed similar efficiency challenges using Mixture of Experts (MoE) architectures, this approach has not yet been adapted for text embeddings.

In this work, we introduce the first general-purpose Mixture of Experts text embedding model.
We demonstrate that scaling text embedding models with Mixture of Experts in both monolingual and multilingual settings outperforms existing approaches while using fewer active parameters.

\begin{table*}
    \centering
    \caption{Evaluation of Multilingual Text Embedding Models}
    \vskip 0.15in
    \begin{tabular}{lrrrrrrr}
        \toprule
        Model & Params (M) & Emb Dim & BEIR & MIRACL & Pretrain Data & Finetune Data & Code \\
        \midrule
        mE5 Base & 278 & 768 & 48.88 & 62.30 & No & No & No \\
        mGTE Base & 305 & 768 & 51.10 & 63.40 & No & No & No \\ 
        Arctic Embed v2 Base & 305 & 768 & \textbf{55.40} & 59.90 & No & No & No \\
        Nomic Embed v2 & 305 & 768 & 52.86 & \textbf{65.80} & \textbf{Yes} & \textbf{Yes} & \textbf{Yes} \\
        \midrule
        BGE M3 & 568 & 1024 & 48.80 & \textbf{69.20} & No & \textbf{Yes} & No \\
        Arctic Embed v2 Large & 568 & 1024 & \textbf{55.65} & 66.00 & No & No & No \\
        mE5 Large & 560 & 1024 & 51.40 & 66.50 & No & No & No \\
        mE5 Large Instruct & 560 & 1024 & 52.64 & 65.70 & No & No & No \\
        Jina Embed v3 & 572 & 1024 & 53.88 & 61.20 & No & No & No \\
        \bottomrule
    \end{tabular}%
    \label{tab:model_comparison}
\end{table*}
\section{Related Work}
\subsection{Mixture of Experts}
The Mixture of Experts (MoE) architecture was first introduced by \citet{shazeer2017outrageouslylargeneuralnetworks} as a method to increase model capacity and performance without a proportional increase in computation by stacking sparsely gated LSTM blocks \citep{hochreiter1997lstm}.
\citet{lepikhin2020gshardscalinggiantmodels} utilized MoE layers in Transformers for machine translation and showed improvements in multilingual translation as the model size increased, while only incurring a sublinear increase in training time.
\citet{fedus2022switchtransformersscalingtrillion} simplified the routing, reduced training instability, and reduced communication costs to achieve a 7x improvement in pre-training speed.
\citet{zoph2022stmoedesigningstabletransferable} found that MoEs frequently experienced training instabilities, and introduced an auxiliary loss to stabilize the model training without harming its quality.

Recent advances in MoE training, such as upcycling from pretrained transformers \citep{komatsuzaki2023sparseupcyclingtrainingmixtureofexperts} and efficient block-sparse implementations \citep{gale2022megablocksefficientsparsetraining}, have made MoE training even more efficient. However, these advances have primarily focused on language modeling tasks. While \citet{hallee2024contrastivelearningmixtureexperts} explored domain-specific MoE embeddings and \citet{li2024mixtureofexpertsllmsecretlyembedding} investigated using MoE language model states as embeddings, our work is the first to develop a general-purpose MoE architecture specifically for text embeddings. Concurrent work GRITLM \citep{muennighoff2024generativerepresentationalinstructiontuning} demonstrates that MoE models like Mixtral 8x7B can effectively handle both embedding and generation tasks through instruction tuning. In contrast, our work focuses on optimizing MoE architectures for embedding efficiency through large-scale contrastive pretraining and finetuning.

\subsection{Monolingual Text Embeddings}
Modern monolingual text embedders typically follow a two-stage approach: contrastive pretraining on large weakly-supervised datasets, followed by contrastive finetuning on human-labeled data \citep{wang2022text,li2023general,günther2023jina,nussbaum2024nomicembedtrainingreproducible}.
Recent work has focused on scaling and data curation \citep{xiao2023cpack,wang2022text,li2023general,günther2023jina,nussbaum2024nomicembedtrainingreproducible,merrick2024arcticembedscalableefficientaccurate,yu2024arcticembed20multilingualretrieval} or adapting decoder-only LLMs for embedding tasks \citep{wang2023improving, lee2024nvembedimprovedtechniquestraining}.

\subsection{Multilingual Text Embeddings}
While multilingual encoders like mBert \citep{devlin2019bert} and XLM-Roberta \citep{conneau2020unsupervisedcrosslingualrepresentationlearning} provide a foundation for cross-lingual representation, they require additional training for high-quality sentence embeddings. Current approaches either rely on translation data \citep{reimers2020makingmonolingualsentenceembeddings} or scale up model size \citep{wang2024multilinguale5textembeddings, chen2024bgem3embeddingmultilingualmultifunctionality}, typically requiring 3-5x more parameters than monolingual models to achieve comparable English performance - a phenomenon known as the ``curse of multilinguality."

Recent work like Arctic Embed 2.0 \citep{yu2024arcticembed20multilingualretrieval} demonstrates that multilingual models can achieve strong English performance without compromising multilingual capability. However, existing approaches still face fundamental challenges with efficiency: state-of-the-art models require large parameter counts and generate large embedding vectors, increasing both computational and economic costs of dense retrieval.

Our MoE-based approach directly addresses this efficiency challenge, maintaining strong performance across both English and multilingual tasks while significantly reducing the active parameter count during inference. This represents a fundamental shift from previous scaling approaches that relied solely on increasing dense model capacity.


\begin{table}[h]
\centering
\caption{MLM Hyperparameters}
\vspace{1.5pt}
\begin{tabular}{ll}
\hline
\textbf{Hyperparameter} & \textbf{Value} \\
\hline
Batch Size & 4,096 \\
Peak Learning Rate & 4e-4 \\
Warmup Steps & 500 \\
Total Steps & 10,000 \\
Grad. Accumulation Steps & 8 \\
Learning Rate Schedule & Linear \\
Sequence Length & 2,048 \\
Rotary Base & 10,000 \\
MLM Probability & 0.3 \\
Language Sampling $\alpha$ & 0.3 \\ 
Max Grad Norm & 1.0 \\
\hline
\end{tabular}
\label{tab:mlm_hyperparameters}
\end{table}

\begin{table*}
\caption{Hyperparameters used for finetuning all models on GLUE benchmark tasks. For mGTE, warmup percentage is set to 6\% and max gradient norm to 1.}
\vskip 0.15in
\centering
\resizebox{\textwidth}{!}{
\begin{tabular}{lcccc|cccccccc}
\toprule
&&&&& \multicolumn{2}{c}{\bf Single Sentence} & \multicolumn{3}{c}{\bf Paraphrase and Similarity} & \multicolumn{3}{c}{\bf Natural Language Inference} \\
\textbf{Model} & \textbf{Params} & \bf Pos. & \bf Seq. & \bf Avg. & CoLA & SST-2 &  MRPC & STS-B & QQP & MNLI & QNLI & RTE \\
\midrule
\midrule
XLM-R-Base & 279M & Abs.  & 512  & 82.35 & 46.95 & 92.54 & 87.37 & 89.32 & 90.69 & 84.34 & 90.35 & 77.26 \\
mNomic-BERT  & 279M & RoPE  & 2048 & 81.63 & 44.69 & 91.97 & 87.50 & 88.48 & 90.93 & 83.59 & 89.38 & 76.54   \\
mGTE-Base & 306M & RoPE & 8192 & 80.77 & 27.22 & 91.97 & 89.71 & 89.55 & 91.20 & 85.16 & 90.91 & 80.41  \\
\bottomrule
\end{tabular}
}
\label{tab:glue-compare}
\end{table*}

\section{Background}
\subsection{Masked Language Modeling}
Masked language modeling (MLM), a self-supervised pretraining objective introduced by \citet{devlin2019bert}, trains a model to recover masked tokens from input sequences.
MLM was applied to both monolingual and multilingual datasets resulting in BERT and mBERT, with the latter demonstrating the potential of cross-lingual representation learning. However, \citet{conneau2020unsupervisedcrosslingualrepresentationlearning} identified that these models were undertrained and introduced XLM-RoBERTa, which achieved performance comparable to monolingual models by training on CC100, a diverse dataset spanning 100 languages from CommonCrawl.

\subsection{Mixture of Experts (MoE)}
Dense models activate all parameters for every input. In contrast, Sparse Mixture of Experts (MoE) models activate only a subset of parameters for each input, reducing computational requirements while maintaining model capacity \citep{shazeer2017outrageouslylargeneuralnetworks}.

In MoE architectures, standard MLP layers are replaced with MoE blocks consisting of multiple ``expert" networks and a router. The router dynamically assigns each input token to a subset of experts using Top-K routing: the router outputs logits for all experts, applies softmax normalization, and routes each token to the $top_k$ experts with the highest probabilities \citep{fedus2022switchtransformersscalingtrillion}.

A key challenge in training MoE models is expert collapse, where certain experts receive disproportionate traffic and others remain underutilized. This is typically addressed through an auxiliary load balancing loss \citep{zoph2022stmoedesigningstabletransferable}:

\begin{equation}
\label{eq:lbl}
    \mathcal{L}_{balance} = \alpha \sum_{i=1}^E (r_i \cdot p_i)
\end{equation}

where $r_i$ is the fraction of tokens routed to expert $i$ and $p_i$ is the mean routing probability for that expert across a batch of tokens. The coefficient $\alpha$ controls the strength of the balancing loss relative to the main objective.

\subsection{Contrastive Learning}
\subsubsection{Training Text Embedding Models}
Text embedding models are generally trained in two stages: weakly-supervised contrastive pretraining and contrastive finetuning \citep{reimers2019sentencebert}. 

The contrastive pretraining stage uses the InfoNCE objective \citep{oord2019representation} to train a biencoder to distinguish relevant text pairs from irrelevant pairs. Given a batch $B = (q_0, d_0), (q_1, d_1) ... (q_n, d_n)$, the objective is:
\begin{equation}\label{eq:infonce}
\mathcal{L}_{C} = -\frac{1}{n} \sum_{i} \log \frac{e^{s(q_i, d_i) / \tau}}{e^{s(q_i, d_i) / \tau} + \sum_{j \neq i}^{n} e^{s(q_i, d_{j}) / \tau}}
\end{equation}
where $s(q, d)$ is the learned score between query $q$ and document $d$ and $\tau$ is the temperature. 
Contrastive finetuning incorporates high-quality human labeled datasets and hard negatives to improve retrieval performance \citep{wang2022text}. The InfoNCE objective is adapted to include these hard negatives:
\begin{align}
Z_i &= e^{s(q_i, d_i) / \tau} + \sum_{j \neq i}^{n} e^{s(q_i, d_{j}) / \tau} + \sum_{m=1}^{H} e^{s(q_i, d_{hn}(1, m)) / \tau} \\
\mathcal{L}_{C} &= -\frac{1}{n} \sum_{i} \log \frac{e^{s(q_i, d_i) / \tau}}{Z_i}
\end{align}

To reduce the storage costs of embedding vectors, which scale with embedding dimension, recent works have applied Matryoshka Representation Learning \citep{kusupati2024matryoshkarepresentationlearning} during both training stages \citep{lee2024geckoversatiletextembeddings}. 
This enables more efficient storage of the computed embeddings by encouraging a rank ordering over the information content of successive embedding subspaces

\subsubsection{Consistency Filtering}
Consistency filtering improves dataset quality by removing potential false positives from weakly supervised data \citep{wang2022text}.
In this approach, each dataset is divided into shards of 1-3M samples.
An existing text embedding model first embeds all queries and documents. Query-document pairs are then discarded if a ground truth document does not appear among the top-k most similar documents to query.

Initially developed for English text embeddings \citep{günther2024jina, nussbaum2024nomicembedtrainingreproducible}, consistency filtering has been adapted for multilingual data by \citet{yu2024arcticembed20multilingualretrieval} using multilingual-E5-small \citep{wang2024multilinguale5textembeddings} with 3M samples per shard and a top-20 filtering threshold.

\begin{table*}
\caption{XTREME-R Benchmark}
\vskip 0.15in
\centering
\resizebox{\textwidth}{!}{
\begin{tabular}{lc|ccccccccccc}
\toprule
\textbf{Model} & \textbf{Avg.} \\
& & XNLI & XCOPA & UDPOS & WikiANN & XQuAD & MLQA & TyDiQA-GoldP & Mewsli-X & LAReQA & Tatoeba \\
& & Acc. & Acc. & F1 & F1 & F1 & F1 & F1 & mAP@20 & mAP@20 & Acc. \\
\midrule
\midrule
XLM-R-Base & 62.31 & 74.49 & 51.8 & 74.33 & 60.99 & 72.96 & 61.45 & 54.31 & 42.45 & 63.49 & 66.79 \\
mNomic-BERT & 62.70 & 73.57 & 61.71 & 74.92 & 60.96 & 71.13 & 59.61 & 43.46 & 43.27 & 67.49 & 70.82   \\
mGTE-Base & 64.63 & 73.58 & 63.62 & 73.52 & 60.72 & 74.71 & 63.88 & 49.68 & 44.58 & 71.90 & 70.07 \\
\bottomrule
\end{tabular}
}
\label{tab:xtremer-compare}
\end{table*}

\subsubsection{Hard Negative Mining}
Text embedding models are typically finetuned with hard negatives mined by an existing retriever \citep{nussbaum2024nomicembedtrainingreproducible, yu2024arcticembed20multilingualretrieval}. While traditional approaches use the top-k most similar documents as hard negatives, this can introduce false negatives. To address this, \citet{moreira2024nv} introduced positive-aware hard negative mining:

\vspace{-5pt}
\begin{equation}
\label{eq:top_k_margin}
threshold = pos\_sim * percentage\_margin
\end{equation}

\vspace{-2pt}
where $percentage\_margin$ (typically 95\%) creates a threshold below which negatives are accepted, reducing false negatives. Recent work has shown that using stronger teacher models for mining yields higher quality finetuning datasets \citep{moreira2024nv, yu2024arcticembed20multilingualretrieval}.


\section{Methods}  
\subsection{Adapting XLM-Roberta for Long-Context}
To extend document-level capabilities to multilingual settings, we modify XLM-Roberta Base \citep{conneau2020unsupervisedcrosslingualrepresentationlearning} to handle longer sequences as XLM-Roberta's absolute positional encodings restrict inputs to 512 tokens. 

Following \citet{gumma2024inducingdocumentlevelabilitiesstandard}, we replace the absolute positional encodings with Rotary Positional Embeddings (RoPE) \citep{su2023roformer}. We set the RoPE base parameter to 10,000, enabling the model to extrapolate to longer sequences while maintaining stable performance. While recent work \citep{liu2024scalinglawsropebasedextrapolation,xiong2023effectivelongcontextscalingfoundation} suggests using larger RoPE bases, our experiments showed degraded performance on GLUE and XTREME-R benchmarks with larger values. This difference might stem from our training approach – unlike \citet{zhang2024mgtegeneralizedlongcontexttext}, who first train with shorter sequences (2,048 tokens) before scaling up, we maintain consistent sequence lengths throughout training.

We use 2048-token segments from a reconstructed CC100 dataset \footnote{https://huggingface.co/datasets/statmt/cc100}. Following the original XLM-Roberta training protocol, we set the language sampling temperature to 0.3. We train for 10,000 steps with hyperparameters detailed in Table \ref{tab:mlm_hyperparameters}. 

We refer to our adapted model as mNomic-BERT.

\subsection{Consistency Filtering}
To ensure high-quality training data, we implement retrieval-based consistency filtering on our multilingual corpus consisting of data from mC4 and multilingual CC News. This approach, established in recent work \citep{yu2024arcticembed20multilingualretrieval, nussbaum2024nomicembedtrainingreproducible}, helps eliminate low-quality or misaligned text pairs from the training set.

For each language in our corpus, we divide the dataset into segments of 1 million examples. Using the multilingual E5 small embedding model \citep{wang2024multilinguale5textembeddings}, we compute similarity between query-document pairs. We retain only pairs where the document ranks among the top 2 most similar documents for its corresponding query, following similar filtering approaches in \citep{wang2022text, günther2023jina}. For English-language data, we utilize the pre-filtered dataset from \citet{nussbaum2024nomicembedtrainingreproducible}.

This filtering process yields a final training dataset of 1.6 billion high-quality pairs. The distribution of data across different languages is detailed in Appendix \ref{appendix:distribution}.

\subsection{Weakly-Supervised Contrastive Pretraining}
\label{sec:pretrain}
For our contrastive pretraining phase, we initialize a biencoder with mNomic-BERT and train it on our filtered contrastive dataset for one epoch. Following \citet{komatsuzaki2023sparseupcyclingtrainingmixtureofexperts}, we transform every alternate MLP layer into an MoE layer with 8 experts and top-2 routing, starting from the second layer. This results in a model with 475M total parameters, of which only 305M are active during inference. We set the load balancing loss coefficient $\alpha$ from Equation \ref{eq:lbl} to 1.

For training, we use the InfoNCE contrastive loss \citep{oord2019representation} with a temperature of $\tau = 0.02$. Following recent work \citep{nussbaum2024nomicembedtrainingreproducible, merrick2024arcticembedscalableefficientaccurate}, we process one dataset per batch with a batch size of 16,384, using random batch sampling. Similar to \citet{yu2024arcticembed20multilingualretrieval}, we set maximum sequence lengths of 32 and 256 tokens for queries and documents respectively due to computational constraints.

We train the model using 16 H100 GPUs with distributed data-parallel training and activation checkpointing. Our optimization uses a peak learning rate of 8e-5 with 1,000 warmup steps and cosine decay.

\subsection{Hard Negative Mining}
For each query in our dataset, we mine hard negatives using a margin-based approach defined in Equation \ref{eq:top_k_margin}. We use the data from \citet{chen2024bgem3embeddingmultilingualmultifunctionality} and BGE M3 for filtering both English and multilingual data. 

\begin{table*}[t]
    \centering
    \caption{MIRACL Performance Across Different Languages. Numbers for E5 taken from \citet{wang2024multilinguale5textembeddings}.} 
    \vskip 0.15in
    \small
    \setlength{\tabcolsep}{2pt}
    \begin{tabular}{l cc *{18}{c}}
    \toprule
    Model & \multicolumn{1}{c}{Avg (18)} & \multicolumn{1}{c}{Avg (16)} & \multicolumn{18}{c}{NDCG@10 Per Language} \\
    \cmidrule(lr){2-21}
     & & & ar & bn & de & en & es & fa & fi & fr & hi & id & ja & ko & ru & sw & te & th & yo & zh \\
    \midrule
    Arctic M v2.0 & 60.6 & 59.9 & 69.7 & 67.7 & \textbf{56.7} & \textbf{55.7} & 55.4 & 52.6 & 68.4 & 54.0 & 53.7 & 48.3 & 58.3 & 59.7 & 58.8 & 52.3 & 81.7 & 74.3 & 75.6 & 48.3 \\
    mGTE Base & 63.6 & 63.4 & 71.4 & 72.9 & 49.7 & 54.0 & 51.8 & 54.0 & 73.5 & 54.5 & 51.9 & 50.3 & 65.8 & 62.9 & 63.2 & 69.9 & \textbf{83.1} & 74.0 & 79.3 & \textbf{61.8} \\
    mE5 Base & 62.2 & 62.3 & 71.6 & 70.2 & 51.9 & 51.2 & 51.5 & 57.4 & 74.4 & 49.7 & 58.4 & 51.1 & 64.7 & 62.2 & 61.5 & \textbf{71.1} & 75.2 & 75.2 & 70.7 & 51.5 \\
    Nomic Embed v2 & \textbf{66.0} & \textbf{65.8} & \textbf{76.7} & \textbf{73.6} & 56.6 & 54.7 & \textbf{56.3} & \textbf{59.2} & \textbf{77.1} & \textbf{55.8} & \textbf{60.5} & \textbf{54.2} & \textbf{67.0} & \textbf{65.9} & \textbf{65.2} & 66.3 & 82.6 & \textbf{78.3} & \textbf{78.3} & 59.5 \\
    \midrule
    Arctic L v2.0 & 66.3 & 66.0 & 76.1 & 74.4 & \textbf{58.6} & 53.7 & 55.6 & 60.3 & 77.1 & 56.7 & 58.4 & 52.3 & 66.5 & 66.3 & 67.1 & 70.8 & 83.5 & 77.5 & 78.3 & 59.9 \\
    mE5 Large & 66.6 & 66.5 & 76.0 & 75.9 & 56.4 & 52.9 & 52.9 & 59.0 & 77.8 & 54.5 & \textbf{62.0} & 52.9 & 70.6 & 66.5 & 67.4 & 74.9 & \textbf{84.6} & 80.2 & 78.3 & 56.0 \\
    E5 Large Instr. & 66.1 & 65.7 & 76.8 & 73.9 & 55.7  & 51.5 & 53.7 & 59.4 & 77.3 & 53.7 & 60.3 & 52.1 & 69.0 & 65.3 & 67.9 & 72.5 & 83.4 & 78.6 & 81.6 & 56.2 \\
    BGE M3 & \textbf{69.2} & \textbf{69.2} & \textbf{78.5} & \textbf{79.9} & 56.8 & \textbf{56.9} & 56.1 & \textbf{60.9} & \textbf{78.6} & \textbf{58.2} & 59.5 & \textbf{56.0} & \textbf{72.8} & \textbf{69.6} & \textbf{70.1} & \textbf{78.6} & \textbf{86.2} & \textbf{82.6} & \textbf{81.8} & \textbf{62.6} \\
    \bottomrule
    \end{tabular}
    \label{tab:miracl_lang_performance}
\end{table*}

\subsection{Contrastive Finetuning}
\label{sec:finetune}
We finetune the pretrained biencoder from Section \ref{sec:pretrain} using our mined hard negatives. For each query, we incorporate 10 hard negative examples during training. We train for one epoch using a batch size of 256, with a peak learning rate of 2e-5, 400 warmup steps, and linear decay. Compared to pretraining, we increase both query and document maximum lengths to 512 tokens.

To enable efficient inference at multiple dimensions, we incorporate Matryoshka Representation Learning \citep{kusupati2024matryoshkarepresentationlearning}, training the model to produce effective embeddings at dimensions 768 and 256. The distribution of our finetuning data is detailed in Appendix \ref{appendix:finetune}.

We refer to this final model as Nomic Embed v2.



\section{Experimental Setup}
\subsection{GLUE Evaluation Protocol}
We evaluate mNomic-BERT on the GLUE benchmark \citep{wang2019glue}, following the evaluation protocol from \citet{nussbaum2024nomicembedtrainingreproducible}. We train each model on 8 GLUE tasks for 3 epochs across 5 random seeds, varying batch sizes (16, 32) and learning rates (1e-5, 2e-5, 3e-5). For mGTE evaluation, we modify these parameters to use 6\% warmup and max gradient norm of 1, matching \citet{zhang2024mgtegeneralizedlongcontexttext}. Following standard practice \citep{liu2019roberta}, we initialize RTE, STSB, and MRPC tasks from an MNLI checkpoint. Table \ref{tab:glue_hyperparameters} details the complete hyperparameter configuration.

\subsection{XTREME-R Evaluation Setup}
We evaluate mNomic-BERT on XTREME-R \citep{ruder2021xtremerchallengingnuancedmultilingual}, a comprehensive benchmark consisting of 10 tasks designed to assess multilingual natural language understanding capabilities. All experiments follow a zero-shot cross-lingual transfer protocol: models are trained exclusively on English data and evaluated on multilingual and cross-lingual tasks. We utilize the evaluation pipeline from \citet{zhang2024mgtegeneralizedlongcontexttext}\footnote{\url{https://github.com/izhx/nlu-evals}} to ensure fair comparison with baseline models XLM-R-Base and mGTE-Base.

\subsection{Text Embedding Benchmark Setup}
We evaluate our model on two retrieval benchmarks: (1) BEIR, the retrieval subset of MTEB \citep{muennighoff2023mteb}, which focuses on English-only retrieval, and (2) MIRACL \citep{zhang2022makingmiraclmultilingualinformation}, which evaluates multilingual retrieval capabilities. For all experiments, we:

\begin{itemize}\setlength{\itemsep}{-0.2em}
    \item Prepend task-specific prefixes ``search\_query" and ``search\_document" to queries and documents
    \item Truncate all inputs to 512 tokens
    \item Measure performance using nDCG@10
\end{itemize}
\vspace{-0.5em}

For reproducibility, we conduct all evaluations using the FlagEmbedding framework\footnote{\href{https://github.com/FlagOpen/FlagEmbedding}{https://github.com/FlagOpen/FlagEmbedding}}, except for mE5 results which are taken directly from \citet{wang2024multilinguale5textembeddings}. Note that mE5 results for German (de) and Yoruba (yo) languages were not reported in the original paper.

\section{Results}
\subsection{mNomic-BERT GLUE Results}
\label{sec:glue-results}
Our approach achieves strong performance across the GLUE benchmark, as shown in Table \ref{tab:glue-compare}. Specifically, mNomic-BERT achieves comparable performance to XLM-R-Base across all tasks, demonstrating that our RoPE-based positional encoding modification and lightweight finetuning preserve the model's capabilities. Notably, mNomic-BERT matches mGTE-Base performance while requiring only 3\% of mGTE-Base's pretraining steps, suggesting that our lightweight finetuning approach effectively extends context length without extensive pretraining.

While \citet{zhang2024mgtegeneralizedlongcontexttext} reported lower CoLA scores for XLM-Roberta, our hyperparameter search revealed that this task is particularly sensitive to configuration choices. We successfully reproduced mGTE-Base's reported CoLA performance but found significant variance across different hyperparameter settings, resulting in a lower median score.

\begin{table}[h]
\centering
\caption{GLUE Fintuning Hyperparameters}
\vskip 0.15in
\begin{tabular}{ll}
\hline
\textbf{Hyperparameter} & \textbf{Value} \\
\hline
\small
Epochs & 3 \\
Sequence Length & 128 \\
Batch Size & 16, 32 \\
Learning Rate & {1, 2, 3}e-5 \\
Learning Rate Schedule & Linear \\
Warmup Pct & 0 \\
Max Grad Norm & 0 \\
\hline
\end{tabular}
\label{tab:glue_hyperparameters}
\end{table}

\subsection{XTREME-R Results}
Table \ref{tab:xtremer-compare} presents the performance of mNomic-BERT compared to XLM-R-Base and mGTE-Base across XTREME-R tasks. mNomic-BERT achieves an average score of 62.70, which is comparable to XLM-R-Base's 62.31 but falls slightly behind mGTE-Base's 64.63. This pattern is consistent across most individual tasks, with mNomic-BERT and XLM-R-Base showing similar performance levels. These results suggest that our approach maintains the cross-lingual capabilities of the base architecture while extending the context length of multilingual text encoders, complementing recent work by \citet{gumma2024inducingdocumentlevelabilitiesstandard}.

\subsection{Text Embedding Benchmark}
We evaluate performance on BEIR, the retrieval subset of MTEB \citep{muennighoff2023mteb}, an English-only benchmark, and MIRACL \citep{zhang2022makingmiraclmultilingualinformation}, a multilingual retrieval benchmark. Results can be found in Table \ref{tab:model_comparison} and \ref{tab:miracl_lang_performance}.

Compared to similarly sized parameter models, Nomic Embed v2 outperforms all models on BEIR and MIRACL except Arctic Embed v2 Base. However, \citet{yu2024arcticembed20multilingualretrieval} do not release any of their training data of which a large percentage consists of private web search data. 

Despite being 2x smaller, Nomic Embed v2 outperforms all multilingual models on BEIR, except Arctic Embed v2 Large, and is competitive with all models on MIRACL.


\section{Analysis}

\subsection{Effectiveness of MoEs for Text Embeddings}
We compare monolingual MoE and dense text embedding models by pretraining them on 235M weakly-supervised contrastive pairs from \citet{nussbaum2024nomicembedtrainingreproducible}. For evaluation, we use the BEIR benchmark \citep{thakur2021beir} across varying batch sizes, with a fixed maximum sequence length of 128 tokens. Our MoE model (Nomic BERT MoE) is created by upcycling alternate layers of Nomic BERT following \citet{komatsuzaki2023sparseupcyclingtrainingmixtureofexperts}. The model uses token choice routing with TopK Routing ($\mathbf{k=1}$, also known as Switch Routing \cite{fedus2022switchtransformersscalingtrillion}) and 8 experts. We compare this against two baselines: the original Nomic BERT and BERT Large \citep{devlin2019bert}.

\begin{figure}[t]
    \centering
    \includegraphics[width=0.95\columnwidth]{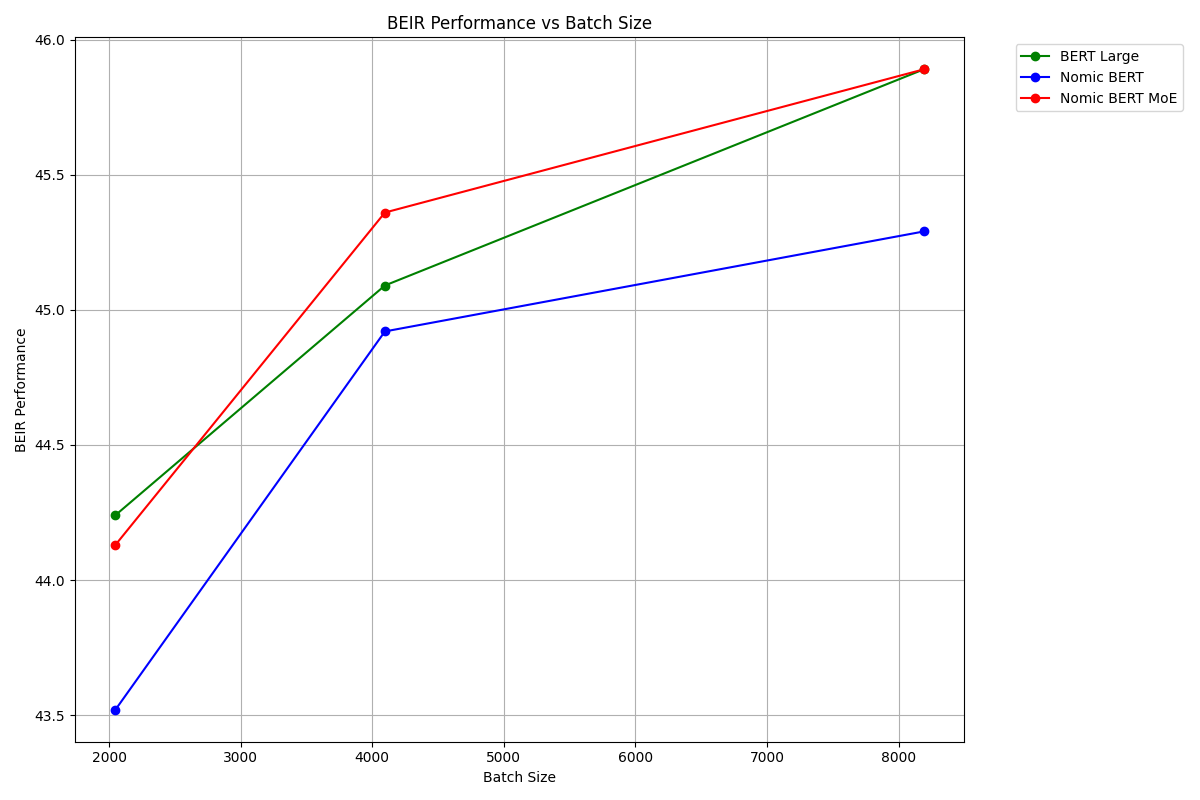}
    \caption{\textbf{Impact of Model Size and Batch Size on Retrieval Performance}. NDCG@10 scores on BEIR benchmark across different batch sizes and model architectures. The upcycled MoE model's performance approaches that of a model with 3x more active parameters as batch size increases, demonstrating efficient scaling behavior.}
    \label{fig:mono_upcycle}
\end{figure}

Figure \ref{fig:mono_upcycle} shows that Nomic BERT MoE consistently outperforms the original Nomic BERT across all batch sizes, despite maintaining a similar number of active parameters. Notably, our MoE model achieves comparable performance to BERT Large, despite the latter having 3x more active parameters, demonstrating the efficiency of the MoE architecture.
 
\begin{table}[h]
\centering
\caption{\textbf{Impact of Upcycled Layers on Model Performance.} BEIR scores across batch sizes and upcycled layers. 6-layer models outperform 12-layer variants at larger batches, suggesting selective upcycling is more effective than full model conversion.}
\vskip 0.15in
\small
\begin{tabular}{ccc}
    \toprule
    MoE Layers & Batch Size & BEIR \\
    \midrule
    \multirow{3}{*}{6} & 2048 & 44.13 \\
     & 4096 & \textbf{45.36} \\
     & 8192 & \textbf{45.89} \\
    \midrule
    \multirow{3}{*}{12} & 2048 & 44.28 \\
     & 4096 & 44.89 \\ 
     & 8192 & 45.48 \\
    \bottomrule
\end{tabular}
\label{tab:upcycle-layers}
\end{table}

Table \ref{tab:upcycle-layers} presents an ablation study on the number of upcycled layers. Converting all 12 layers to MoE layers actually reduces performance compared to converting only 6 layers, particularly at larger batch sizes. This suggests that selective layer upcycling provides a better balance between model capacity and optimization stability.

\subsection{Effectiveness of MoEs for Multilingual Text Embeddings}
We extend our analysis to the multilingual setting by incorporating an additional 65M weakly-supervised contrastive pairs from mC4 \citep{xue2021mt5massivelymultilingualpretrained} and Multilingual CC News \citep{wang2024multilinguale5textembeddings}. For a controlled ablation study, we focus on six languages spanning different language families: English, Chinese, Arabic, Hindi, Spanish, and Swahili. This selection includes both high-resource and low-resource languages, with Swahili representing the latter category. We evaluate three models: XLM-RoBERTa Base \citep{conneau2020unsupervisedcrosslingualrepresentationlearning}, our MoE variant (XLM-RoBERTa MoE Base), and XLM-RoBERTa Large. Performance is measured using NDCG@10 on both BEIR \citep{thakur2021beir} and MIRACL \citep{zhang2022makingmiraclmultilingualinformation} benchmarks across different batch sizes.

Table \ref{tab:multilingual-model-comparison} presents our multilingual evaluation results. While our MoE model consistently outperforms its dense counterpart across all batch sizes on both BEIR and MIRACL benchmarks, it does not match the performance of the larger model—a notable departure from our monolingual findings.

Our experiments reveal that data scale significantly impacts the performance of XLM-RoBERTa MoE Base. Initial experiments with a smaller dataset of 100M total contrastive pairs showed the MoE model consistently underperforming its parameter-equivalent dense counterpart. This aligns with findings from \citet{krajewski2024scalinglawsfinegrainedmixture}, who observed that MoE models tend to underperform dense models under limited training regimes.

\begin{table}[htbp]
    \centering
    \caption{Evaluation of different teacher models and thresholds for hard negative mining}
    \vskip 0.15in
    \resizebox{\columnwidth}{!}{%
    \begin{tabular}{lrrrrrr}
        \toprule
        Teacher Model & Margin & Negatives & Avg & FiQA & HotpotQA & NQ \\
        \midrule
        Arctic Embed Large & None & 4 & 52.87 & 42.68 & 59.47 & 56.45 \\
        Arctic Embed Large & 0.95 & 4 & 55.20 & 44.98 & 62.75 & 57.88 \\
        NVEmbed v1 & 0.95 & 4 & 54.94 & 45.00 & 58.29 & 61.56 \\
        Stella 1.5B & 0.95 & 4 & 57.22 & 45.18 & 64.06 & 62.42 \\
        Stella 1.5B & 0.98 & 4 & 57.20 & 45.67 & 63.65 & 62.29 \\
        Stella 1.5B & 0.95 & 7 & 57.31 & 45.10 & 64.27 & 62.58 \\
        Stella 1.5B & 0.95 & 10 & 57.45 & 45.17 & 64.48 & 62.69 \\
        \bottomrule
    \end{tabular}%
}
    \label{tab:hard-negative-mining-ablation}
\end{table}
\subsection{Hard Negative Mining}
\label{sec:hn_mine}
We investigate the impact of different teacher models and margin thresholds for hard negative mining, following the approach of \citet{moreira2024nv}. We initialize our model from E5-Large Unsupervised \citep{wang2022text}\footnote{\url{https://huggingface.co/intfloat/e5-large-unsupervised}} and mine negatives using Equation \ref{eq:top_k_margin}. Our training data comprises approximately 500k examples from three sources: StackExchange Title-Body pairs\footnote{\url{https://huggingface.co/datasets/sentence-transformers/embedding-training-data}}, SQuAD \citep{rajpurkar-etal-2016-squad}, and Natural Questions (NQ) \citep{kwiatkowski-nq}. We evaluate performance on three BEIR datasets: NQ, FiQA, and HotpotQA \citep{thakur2021beir}. For teacher models, we compare NVEmbed v1 \citep{lee2024nv}, Arctic Embed Large \citep{merrick2024arcticembedscalableefficientaccurate}, and Stella 1.5B v5 \citep{zhang2025jasperstelladistillationsota}.

Table \ref{tab:hard-negative-mining-ablation} presents our findings across different teacher models and mining parameters. Several key trends emerge: Positive aware hard negative mining with consistently improves performance, as shown by the 2.33 point average improvement when using Arctic Embed Large with a margin of 0.95 compared to no margin. Surprisingly, Stella 1.5B outperforms NVEmbed v1 even though it is a 7x smaller model. Increasing the number of negative examples from 4 to 10 with Stella 1.5B yields modest but consistent improvements, with the best average performance of 57.45 achieved using 10 negatives. However, the gains diminish with each additional negative, suggesting a potential plateau in the benefits of increased negative examples. Finally, varying the margin threshold between 0.95 and 0.98 shows minimal impact on overall performance, indicating that the mining process is relatively robust to this hyperparameter within this range.

We also compared our best-performing mined dataset against a filtered version of the finetuning data released by \citet{chen2024bgem3embeddingmultilingualmultifunctionality}. Using BGE M3 to filter negatives based on Equation \ref{eq:top_k_margin}, this approach achieved 1 point higher NDCG@10 on BEIR, suggesting filtering potential negatives from an existing mined dataset is also a viable option.

\begin{table}[htbp]
    \centering
    \small  
    \caption{\textbf{Performance Comparison of Multilingual Models.} BEIR and MIRACL scores across different model architectures and batch sizes. XLM-R Large (561M parameters) consistently outperforms both the MoE variants and the base model (XLM-B, 278M parameters). MoE models show improved performance with increased batch sizes, particularly when using k=2 experts.}
    \vskip 0.15pt
    \begin{tabular}{l@{\hspace{8pt}}r@{\hspace{8pt}}r@{\hspace{8pt}}r}
        \toprule
        \textbf{Model} & \textbf{Params} & \textbf{BEIR} & \textbf{MIRACL} \\
        \midrule
        \multicolumn{4}{l}{\textit{Batch Size: 2048}} \\
        \addlinespace[0.2em]
        XLM-R Large & 561M & \textbf{45.86} & 38.49 \\
        XLM-MoE (k=1) & 278M  & 43.63 & \textbf{39.17} \\
        XLM-B & 278M & 43.51 & 34.14 \\
        \midrule
        \multicolumn{4}{l}{\textit{Batch Size: 4096}} \\
        \addlinespace[0.2em]
        XLM-R Large & 561M & \textbf{46.26} & \textbf{38.99} \\
        XLM-MoE (k=1) & 278M & 44.37 & 37.08 \\
        XLM-B & 278M & 43.78 & 37.32 \\
        \midrule
        \multicolumn{4}{l}{\textit{Batch Size: 8192}} \\
        \addlinespace[0.2em]
        XLM-R Large & 561M & \textbf{46.91} & \textbf{42.71} \\
        XLM-MoE (k=2) & 300M & 45.00 & 39.81 \\
        XLM-MoE (k=1) & 278M & 44.11 & 39.17 \\
        XLM-B & 278M & 43.96 & 37.92 \\
        \bottomrule
    \end{tabular}
    \label{tab:multilingual-model-comparison}
\end{table}

\section{Conclusion}
We introduce Nomic Embed v2, the first Mixture of Expert Embedding Model. Nomic Embed v2 outperforms similarly sized and larger embedding models in both English and Multilingual Retrieval benchmarks while being trained only publicly available data. Nomic Embed v2 proves a successful alternative to scaling text embedding models without increasing computational costs.

\section{Limitations and Future Work}

Our work with Nomic Embed v2 demonstrates the advantages of MoE architectures over dense models for text embeddings. However, this represents only an initial exploration of MoE applications in this domain. Several promising research directions emerge: investigating the optimal scaling of expert count and active parameters, exploring alternative routing mechanisms, and examining how loss-free routing could leverage the bidirectional nature of these models. Furthermore, techniques for distilling MoE models back into dense architectures could make these improvements more widely deployable.

Beyond architectural choices, understanding the fundamental scaling relationships between dataset size, model parameters, and embedding dimension would provide valuable insights for the field. This could help establish whether the benefits of MoE architectures persist or even compound at larger scales.

\newpage
\section*{Impact Statement}
This paper presents work whose goal is to advance the field of 
Machine Learning. There are many potential societal consequences 
of our work, none which we feel must be specifically highlighted here.

\bibliography{multi}
\bibliographystyle{multi}

\newpage
\appendix
\onecolumn
\section{Weakly Supervised Contrastive Pretraining Dataset Distribution}
The full pretraining dataset distribution can be see in Table \ref{tab:full}.

\label{appendix:distribution}
\begin{table}[htbp]
    \centering
    \label{tab:full}
    \caption{Dataset Distribution of 1.6B pairs for weakly supervised contrastive pretraining }
    \small
    \begin{tabular}{llr|llr|llr}
        \toprule
        \textbf{Code} & \textbf{Language} & \textbf{Pairs} & 
        \textbf{Code} & \textbf{Language} & \textbf{Pairs} &
        \textbf{Code} & \textbf{Language} & \textbf{Pairs} \\
        \midrule
        en & English & 234,553,344 & be & Belarusian & 589,824 & my & Burmese & 147,456 \\
        es & Spanish & 210,010,112 & ml & Malayalam & 557,056 & km & Khmer & 131,072 \\
        fr & French & 172,769,280 & kn & Kannada & 524,288 & mg & Malagasy & 131,072 \\
        de & German & 169,426,944 & mk & Macedonian & 425,984 & pa & Punjabi & 131,072 \\
        it & Italian & 104,251,392 & ur & Urdu & 409,600 & ru-Latn & Russian (Latin) & 131,072 \\
        pt & Portuguese & 87,982,080 & fy & Frisian & 393,216 & sn & Shona & 131,072 \\
        pl & Polish & 63,209,472 & fil & Filipino & 360,448 & zh-Latn & Chinese (Latin) & 131,072 \\
        nl & Dutch & 50,118,656 & te & Telugu & 360,448 & ha & Hausa & 98,304 \\
        tr & Turkish & 49,053,696 & eu & Basque & 344,064 & he & Hebrew & 98,304 \\
        ja & Japanese & 43,433,984 & sw & Swahili & 327,680 & hmn & Hmong & 98,304 \\
        vi & Vietnamese & 40,058,880 & so & Somali & 294,912 & ht & Haitian & 98,304 \\
        ru & Russian & 38,731,776 & sd & Sindhi & 262,144 & ja-Latn & Japanese (Latin) & 98,304 \\
        id & Indonesian & 36,470,784 & uz & Uzbek & 262,144 & su & Sundanese & 98,304 \\
        ar & Arabic & 33,800,192 & co & Corsican & 245,760 & bg-Latn & Bulgarian (Latin) & 65,536 \\
        cs & Czech & 29,966,336 & hr & Croatian & 245,760 & gd & Scots Gaelic & 65,536 \\
        ro & Romanian & 24,772,608 & gu & Gujarati & 229,376 & ny & Nyanja & 65,536 \\
        sv & Swedish & 24,608,768 & hi-Latn & Hindi (Latin) & 229,376 & ps & Pashto & 65,536 \\
        el & Greek & 22,773,760 & ceb & Cebuano & 196,608 & ku & Kurdish & 49,152 \\
        uk & Ukrainian & 19,841,024 & eo & Esperanto & 196,608 & sh & Serbo-Croatian & 49,152 \\
        zh & Chinese & 18,661,376 & jv & Javanese & 196,608 & am & Amharic & 32,768 \\
        hu & Hungarian & 18,448,384 & la & Latin & 196,608 & ig & Igbo & 32,768 \\
        da & Danish & 14,548,992 & zu & Zulu & 196,608 & lo & Lao & 32,768 \\
        no & Norwegian & 12,812,288 & mn & Mongolian & 180,224 & mi & Maori & 32,768 \\
        hi & Hindi & 12,713,984 & si & Sinhala & 180,224 & nn & Norwegian Nynorsk & 32,768 \\
        fi & Finnish & 12,697,600 & el-Latn & Greek (Latin) & 163,840 & sm & Samoan & 32,768 \\
        bg & Bulgarian & 12,042,240 & ga & Irish & 163,840 & yi & Yiddish & 32,768 \\
        ko & Korean & 10,354,688 & ky & Kyrgyz & 163,840 & st & Sotho & 16,384 \\
        sk & Slovak & 8,962,048 & tg & Tajik & 163,840 & tl & Tagalog & 16,384 \\
        th & Thai & 7,602,176 & & & & xh & Xhosa & 16,384 \\
        iw & Hebrew & 5,783,552 & & & & yo & Yoruba & 16,384 \\
        ca & Catalan & 5,701,632 & & & & & & \\
        lt & Lithuanian & 5,242,880 & & & & & & \\
        fa & Persian & 5,177,344 & & & & & & \\
        ms & Malay & 4,325,376 & & & & & & \\
        sl & Slovenian & 4,259,840 & & & & & & \\
        lv & Latvian & 3,211,264 & & & & & & \\
        mr & Marathi & 2,588,672 & & & & & & \\
        bn & Bengali & 2,457,600 & & & & & & \\
        sq & Albanian & 2,113,536 & & & & & & \\
        cy & Welsh & 2,048,000 & & & & & & \\
        \bottomrule
    \end{tabular}
    \label{tab:all-language-sizes}
\end{table}

\section{Contrastive Finetuning Dataset Distribution}
Full finetuning data distribution can be found in Table \ref{tab:finetune_ds}. We train on the training sets of BEIR and MIRACL as well as SQuAD and Stackoverflow. 

\section{BEIR Retrieval Performance}
The full BEIR results broken down by task can be found in Table \ref{tab:beir_full}. Nomic Embed v2 at 256 dimensions performs competitively to full dimensionality. 

\label{appendix:finetune}
\begin{table}[!t]
\centering
\label{tab:finetune_ds}
\begin{minipage}{0.48\textwidth}
\centering
\caption{Contrastive finetuning data distribution}
\vskip 0.15pt
\small
\begin{tabular}{l r}
\toprule
\textbf{Dataset} & \textbf{Number of Samples} \\
\midrule
\multicolumn{2}{l}{\textit{English Datasets}} \\
MSMARCO & 485,120 \\
Stack & 249,856 \\
SQuAD & 87,552 \\
HotPot & 82,432 \\
NQ & 57,856 \\
FEVER & 28,672 \\
\midrule
\multicolumn{2}{l}{\textit{MIRACL Train Datasets}} \\
Russian & 4,608 \\
Indonesian & 3,840 \\
Arabic & 3,328 \\
Japanese & 3,328 \\
Telugu & 3,328 \\
Finnish & 2,816 \\
Thai & 2,816 \\
English & 2,560 \\
Spanish & 2,048 \\
Persian & 2,048 \\
Swahili & 1,792 \\
Bengali & 1,536 \\
Chinese & 1,280 \\
French & 1,024 \\
Hindi & 1,024 \\
Korean & 768 \\
\midrule
\textbf{Total} & \textbf{1,029,632} \\
\bottomrule
\end{tabular}
\label{tab:finetune_distribution}
\end{minipage}%
\hfill
\begin{minipage}{0.48\textwidth}
\centering
\label{tab:beir_full}
\caption{BEIR Retrieval Performance}
\vskip 0.15pt
\begin{tabular}{l|cc}
\hline
Dataset & Nomic Embed v2 & Nomic Embed v2 256 \\
\hline
Average & 52.86 & 49.63 \\
ArguAna & 55.73 & 51.97 \\
ClimateFEVER & 33.38 & 29.58 \\
CQADupstack & 42.64 & 40.25 \\
DBPedia & 41.44 & 37.51 \\
FEVER & 87.17 & 84.64 \\
FiQA2018 & 38.74 & 35.71 \\
HotpotQA & 68.53 & 63.67 \\
MSMARCO & 40.89 & 38.90 \\
NFCorpus & 34.60 & 31.29 \\
NQ & 60.41 & 56.12 \\
QuoraRetrieval & 87.95 & 87.49 \\
SCIDOCS & 19.25 & 17.71 \\
SciFact & 72.89 & 66.31 \\
TRECCOVID & 78.78 & 74.22 \\
Touche2020 & 30.55 & 29.03 \\
\hline
\end{tabular}
\label{tab:beir}
\end{minipage}
\end{table}

\end{document}